# A hybrid MLP-PNN architecture for fast image superresolution

Carlos Miravet[1,2] and Francisco B. Rodríguez[1]

[1] Grupo de Neurocomputación Biológica (GNB), Escuela Politécnica Superior,
Universidad Autónoma de Madrid, 28049 Madrid, Spain
carlos.miravet@ii.uam.es, francisco.rodriguez@ii.uam.es
[2] SENER Ingeniería y Sistemas, S. A.,Severo Ochoa 4 (P.T.M.), 28760 Madrid, Spain

**Abstract.** Image superresolution methods process an input image sequence of a scene to obtain a still image with increased resolution. Classical approaches to this problem involve complex iterative minimization procedures, typically with high computational costs. In this paper is proposed a novel algorithm for superresolution that enables a substantial decrease in computer load. First, a probabilistic neural network architecture is used to perform a scattered-point interpolation of the image sequence data. The network kernel function is optimally determined for this problem by a multi-layer perceptron trained on synthetic data. Network parameters dependence on sequence noise level is quantitatively analyzed. This super-sampled image is spatially filtered to correct finite pixel size effects, to yield the final high-resolution estimate. Results on a real outdoor sequence are presented, showing the quality of the proposed method.

## Introduction

Image super-resolution methods process an input image sequence of a scene to obtain a still image with increased resolution. From the earliest algorithm proposed by Tsai and Huang [1], superresolution has attracted a growing interest as a purely computational mean of increasing the imaging sensors performance. During last years, a number of approaches have been proposed [2-7]. Of these, methods based on projection onto convex sets (POCS) [4, 5] and, particularly, bayesian MAP methods [6, 7] have gained acceptance due to their robustness and flexibility to incorporate *a priori* constraints. However, both methods approach the solution by iterative procedures highly demanding in terms of computational cost.

In this paper, we propose a neural network based method that learns from examples how to perform image superresolution with significantly lower computational needs. The method is based on a two-step sequential processing scheme that is applied to the input sequence, increasing the image sampling rate and restoring the degradations associated to the low resolution pixel size.

To operate, the algorithm requires the previous knowledge of the geometrical transformations that relate the input sequence frames. To this end, an adaptation of a classical sub-pixel registration procedure [8, 9] has been used.

After registration, a scattered-point interpolation is performed in first place, using the proposed neural architecture. Finally, this densely sampled image is spatially filtered to restore degradations associated to pixel size and any residual artifacts introduced by the interpolation procedure. Filter coefficient generation is described elsewhere due to space constraints. Application of this second processing step yields the final high-resolution image reconstruction.

In next section, we describe in detail the proposed scattered-point sequence interpolation method, and we analyze quantitatively the network parameters dependence on input sequence noise level. Then, we present experimental results on an outdoor sequence and draw some conclusions in last section.

## Scattered-point interpolation

Once the sequence has been registered to sub-pixel accuracy, all pixels of the input sequence could be conceptually projected onto a common reference system, aligned to the reference sequence frame. This results in a distorted (translated, rotated, scaled) superposition of discrete image grids that represent noisy samples of an underlying continuous image, smoothed by the finite pixel size of the acquiring imaging sensor.

In this step of the process, a scattered-point interpolation is performed on the projected image data to increase the image sampling rate to that of the high resolution image. This step will drastically reduce the aliasing artifacts induced by the finite sampling frequency, improving simultaneously the signal-to-noise ratio of the result due to the weighted averaging implicit in the interpolation operation.

The scattered-point interpolation is performed using a probabilistic neural network (PNN), an architecture introduced by Specht [10]. The PNN is a multivariate kernel density estimator, originally with fixed kernel width. This technique is closely related to a non-parametric regression technique, the Nadaraya-Watson estimator [11], and to probability density estimation methods, such as the Parzen windows [12] method.

By means of this interpolation process, the image values at the nodes of a high-resolution grid are estimated. Each node site is handled independently, which makes this process highly amenable to parallel implementation. For each grid site, the nearest pixel in each projected frame of the N-length input sequence is determined, and its pixel value and distance to the target location is stored as an element of an array of size N. This array constitutes the input to our network.

A diagram of the proposed neural architecture is depicted in figure 1. The first network layer is composed of a set of identical units, one per sequence frame. Each unit determines the relative contribution of the correspondent pixel in the input array to the interpolated value provided by the network. This weight is obtained by applying a non-linear kernel function to the pixel distance to the target location. The second network layer contains summation units that add and normalize all contributions to provide the final interpolated image value.

The selection of a kernel shape and width has a significant impact on the quality of the results. Specht proposed several function kernels, including the exponential, to be considered for application. The kernel width of a PNN is commonly obtained by a trial-and-error procedure. A too narrow kernel will typically lead to spiky, noisy esti-mates of the function to be approximated. On the other hand, a too large kernel width will provide an excessive degree of smoothing, sweeping out the function details.

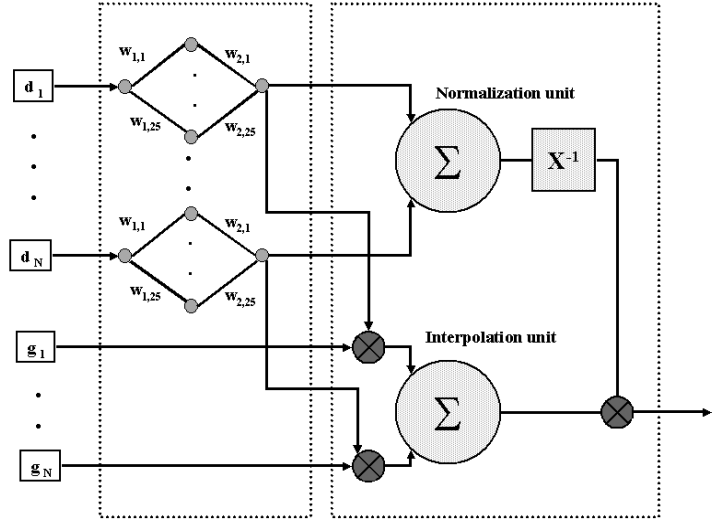

**Fig. 1.** Hybrid MLP-PNN neural architecture for scattered point image sequence interpolation. The network estimates the image value at a target location as a weighted average of the values ($g_1,\ldots, g_N$) of pixels closest to that location in each frame of the N-length sequence. First layer MLP networks compute pixel interpolation weights as a function of distance to the target location. Second layer units perform the weighted average operation to obtain the output value

In our system, the kernel function is determined optimally for the image scattered-point interpolation task by a multi-layer perceptron (MLP), constituting the core of PNN first layer units. The MLP weights, identical for all units, are determined by training the full neural architecture on a set of synthetically generated image sequences, where target values are readily available. The details involved in the generation of these sequences are discussed in the section devoted to network training.

For this purpose, we have used a two-layer perceptron with 25 hidden units and hyperbolic tangent activation functions in the first layer, and a single unit with a linear activation function in the output layer. The network has been trained using a conjugate gradient descent [13] method with gradient computed by an adaptation of the backpropagation algorithm [11] to our neural architecture.

Considering the standard sum-of-squares error function, the error for pattern $k$ is given by:

$$E^k = \frac{1}{2}\left(o^k - t^k\right)^2 \qquad (1)$$

where $o^k$ is the network estimate for the pixel location associated to the $k^{th}$ input array, and $t^k$ is the target image value at that location.

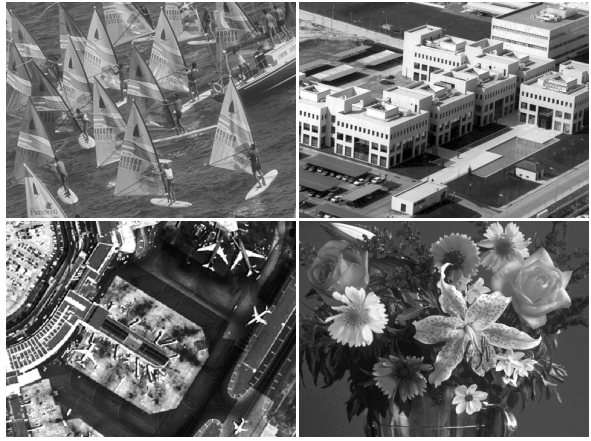

**Fig. 2.** Image set used for synthetic image sequence generation

Using the chain rule, the following expression could be obtained for the derivative of the $k$ pattern error with respect to the output of the $s^{th}$ MLP in our neural architecture:

$$\frac{\partial E^k}{\partial y_s} = \left(o^k - t^k\right) \cdot \frac{\partial o^k}{\partial y_s} = \left(o^k - t^k\right) \cdot \left( \frac{\left(g^k_s \cdot \sum_i y^k_i\right) - \sum_i g^k_i \cdot y^k_i}{\left(\sum_i y^k_i\right)^2} \right) \quad (2)$$

where $y^k_s$ is the output of the $s^{th}$ unit MLP when the network is fed with pattern $k$, and $g^k_s$ is the value of the $s^{th}$ frame nearest pixel, for input pattern $k$. Application of the backpropagation algorithm enables the computation of the weight error gradient in terms of this error derivative with respect to the MLP output.

**Network training**

**Training data generation** Training of the proposed hybrid MLP-PNN architecture has been conducted using a supervised, learning by examples procedure. In this scheme, low-resolution image data, randomly scattered around a target pixel location, is used as input to the network, which predicts the image value at that location. Comparison with the actual pixel value generates an error term which, accumulated over the complete training set, drives the training procedure.

In our approach, training sequence patterns are generated synthetically from input still images. For this purpose, a training still image set composed of four images of widely different image content was used. This image set is presented in figure 2.

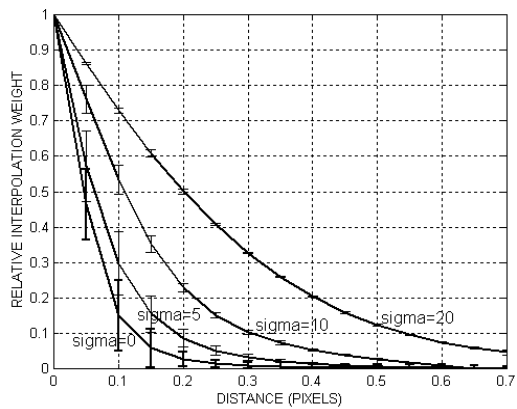

**Fig. 3.** MLP-derived interpolation kernels for different noise variances

To generate a training pattern from a high-resolution still image, a target location on that image has to be previously selected. This location is generated by a probabilistic procedure, where sites in the vicinity of abrupt changes in gray level are assigned a higher probability. The value of a low-resolution pixel centered at this location is computed by adequate bilinear interpolation and averaging operations. This value constitutes the target output of the network. Network input is composed of a set of low-resolution image data randomly scattered around that target location. Gaussian noise can be added to this input data to train the network for different sequence noise levels. Data sets have been generated for different input noise levels, sequence lengths and low-to-high resolution scale factors.

**Training results** The network has been trained on the described synthetic data sets using a conjugate gradient descent method. To avoid local minima, the training procedure was carried out on multiple network realizations with weights randomly initialized, retaining the global minimum. This procedure was validated by comparing the results with those obtained on a subset of cases using a stochastic minimization process based on simulated annealing [14].

In figure 3 are presented the interpolation kernels obtained for training sets with gaussian noise of several standard deviations (0, 5, 10 and 20 gray levels) added to the input low-resolution data. In all cases, the network operates on 25-frame sequences, and provides an output image scaled by a factor of 3 with respect to input sequence frames.

Obtained results show a steady kernel width increase with input data noise level. Intuitively, at low noise levels, preservation of image detail is the primary factor, and kernel widths are adjusted to impose object continuity in the estimated image, avoiding jagged edges typical of sequence nearest-neighbor interpolation. At higher levels, noise becomes a leading factor in the reconstruction error, and kernels widen correspondingly to smooth the results, at the cost of some loss in detail preservation. The training algorithm performs these kernel shape modifications in an autonomous, data-driven manner.

**Table 1.** Interpolation errors for sequence nearest neighbor and MLP-PNN interpolation schemes

| Method | σ=0 | σ=5 | σ=10 | σ=20 |
|---|---|---|---|---|
| SEQ NN | 7.736 | 9.279 | 12.635 | 21.498 |
| MLP-PNN | 4.458 | 5.254 | 6.143 | 7.690 |

In table 1 are presented the obtained interpolation errors for the hybrid MLP-PNN network. Nearest-neighbor sequence interpolation (SEQ NN) has also been included for comparison. In this scheme, the image value at the target location is estimated as the value of the nearest pixel in the projected input sequence.

For zero noise samples, the direct nearest-neighbor scheme provides an interpolation error approximately 70% higher than that of the MLP-PNN network. The difference in the error of both methods increases with the amount of input noise. For input image sequences degraded by additive gaussian noise of 20 gray levels standard deviation, the nearest-neighbor method yields an interpolation error near to 200% higher than that of the corresponding MLP-PNN network.

In a similar way to what has been described so far, preliminary studies were conducted to evaluate interpolation kernel dependence on image content, input sequence length and input/output scale factor. Results, in all cases, reflect a remarkable interpolation kernel independence on these factors, showing the generality of the proposed approach. In a second phase of this study, the effects of other factors, such as misregistration will be analyzed in detail.

## Experimental Results

The proposed algorithm has been tested with excellent performance on sequences with widely different image content. On figure 4 are presented the results obtained on an outdoor image sequence [15] well suited to test superresolution algorithm performance. The input sequence contains 25 frames showing a car gradually moving away from the viewpoint position. Superresolution results on the car license plate zone are presented for a well-known bayesian method [6], a recently proposed method that obtains superresolution by solving a preconditioned Tikhonov regularization problem [16], and our proposed method. As it appears, the perceptual quality of the results obtained with the bayesian method and the method proposed is similar, and somewhat higher than the quality provided by the Tikhonov regularization method. All three methods provided a very significant increase in quality over direct bilinear interpolation of one sequence frame.

Optimized versions of the bayesian and the proposed method were implemented on a Pentium IV commercial PC platform, and their execution time on the license plate zone of the car sequence was measured. The bayesian method required a processing time of 4.47 s to achieve the reported result. This time is almost 150 times higher than that required by the proposed method (0.03 s). The results corresponding to the preconditioned Tikhonov regularization method were taken directly from those reported by the author, and its execution time was not measured on our platform. However, the reported execution times of this iterative method on similar sequences using a Sparc 20 platform are in the range of 10-100 s, which, even after platform processing power compensation, are more than an order of magnitude higher than those provided by the method proposed.

The very significant decrease obtained in processing time requirements, paves the way to quasi-real time applications of superresolution.

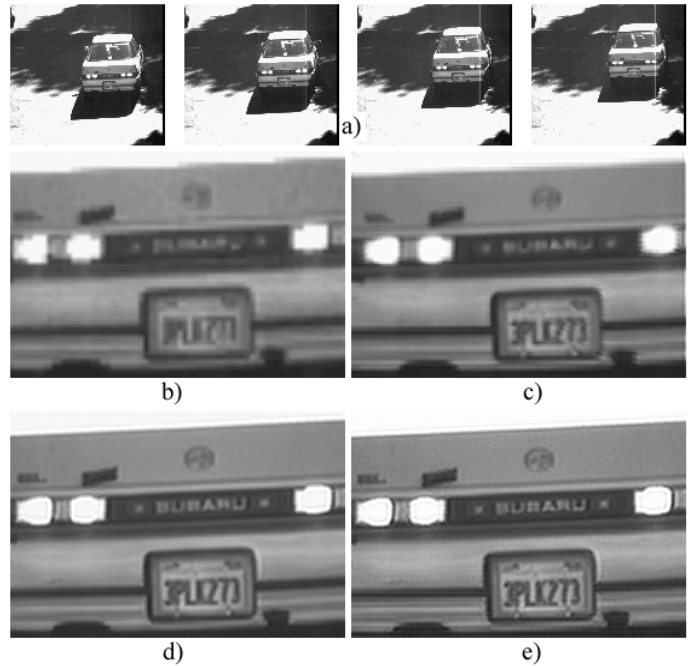

**Fig. 4.** Superresolution results on the "car" sequence. a) Several input sequence frames; b) bi-linear interpolation of the car license plate zone on the central sequence frame; c) results using a preconditioned Tikhonov regularization method (Nguyen *et al*); d) results using a bayesian method (Hardie *et al*); e) results using the hybrid MLP-PNN proposed method. The method proposed provided a measured speed-up factor of about 150 over the execution time of Hardie *et al* method.

## Conclusions

In this paper is proposed a novel, highly-computationally efficient algorithm for image superresolution. The proposed algorithm is based on the sequential application of scattered-point interpolation and restoration operations on the input sequence data. A hybrid MLP-PNN neural network is used to perform the scattered-point interpolation step, where the MLP is trained on synthetically generated sequences to derive optimum interpolation kernel shapes. The restoration process is performed applying an spatial linear filter optimized for this problem. This filter restores the degradations caused by the input low-resolution image pixel size, and reduces any residual artifacts caused by the interpolation procedure. The filter coefficients have been computed for several input sequence noise levels.

The proposed method have been compared with iterative superresolution methods, providing similar or better quality results with a very significant reduction in computational load, paving the way to quasi-real time applications of superresolution techniques.

**Acknowledgments**
This work has been partially supported by TIC2001-0572-C02-02 grant.